\documentclass{article}

\usepackage{PRIMEarxiv}

\usepackage[utf8]{inputenc} 
\usepackage[T1]{fontenc}    
\usepackage{hyperref}       
\usepackage{url}            
\usepackage{booktabs}       
\usepackage{amsfonts}       
\usepackage{nicefrac}       
\usepackage{microtype}      
\usepackage{lipsum}
\usepackage{fancyhdr}       
\usepackage{graphicx}       
\graphicspath{{media/}}     

\usepackage{lipsum}
\usepackage{graphicx}
\graphicspath{ {./images/} }
\usepackage{amsmath}
\usepackage{amssymb}
\usepackage{multirow}
\usepackage{tabularx}
\usepackage{xcolor}
\usepackage{subcaption}
\usepackage{algorithm}
\usepackage{algorithmic}
\usepackage{float}
\usepackage{makecell}

\title{Breaking Forgetting: Training-Free Few-Shot Class-Incremental Learning via Conditional Diffusion}

\author{
  Haidong Kang\thanks{Corresponding author} \\
  Northeastern University \\
  \texttt{hdkustl@gmail.com} \\
  \And
  Ketong Qian \\
  School of Information and Intelligent Science \\
  Donghua University \\
  \texttt{231010206@mail.dhu.edu.cn} \\
  \And
  Yi Lu \\
  Whiting School of Engineering \\
  Johns Hopkins University \\
  \texttt{121090386@link.cuhk.edu.cn} \\
}

\begin{document}
\maketitle
\begin{abstract}
Efforts to overcome catastrophic forgetting in Few-Shot Class-Incremental Learning (FSCIL) have primarily focused on developing more effective gradient-based optimization strategies. In contrast, little attention has been paid to the training cost explosion that inevitably arises as the number of novel classes increases, a consequence of relying on gradient learning even under extreme data scarcity. More critically, since FSCIL typically provides only a few samples for each new class, gradient-based updates not only induce severe catastrophic forgetting on base classes but also hinder adaptation to novel ones. This paper seeks to break this long-standing limitation by asking: Can we design a training-free FSCIL paradigm that entirely removes gradient optimization? We provide an affirmative answer by uncovering an intriguing connection between gradient-based optimization and the Conditional Diffusion process. Building on this observation, we propose a Conditional Diffusion-driven FSCIL (CD-FSCIL) framework that substitutes the conventional gradient update process with a diffusion-based generative transition, enabling training-free incremental adaptation while effectively mitigating forgetting. Furthermore, to enhance representation under few-shot constraints, we introduce a multimodal learning strategy that integrates visual features with natural language descriptions automatically generated by Large Language Models (LLMs). This synergy substantially alleviates the sample scarcity issue and improves generalization across novel classes. Extensive experiments on mainstream FSCIL benchmarks demonstrate that our method not only achieves state-of-the-art performance but also drastically reduces computational and memory overhead, marking a paradigm shift toward training-free continual adaptation.
\end{abstract}


\section{Introduction}

Few-Shot Class-Incremental Learning (FSCIL)~\cite{tao2020few} has emerged as a critical paradigm for real-world continual learning, where models must rapidly incorporate novel classes from only a few labeled examples (e.g., $K{=}5$) while retaining recognition performance on previously learned base classes. Early FSCIL methods, inheriting from general class-incremental learning, relied on replay~\cite{rebuffi2017icarl}, regularization~\cite{kirkpatrick2017ewc, li2017lwf}, or dynamic architectures. However, under extreme data scarcity, the field shifted toward freezing a backbone pre-trained on base classes and adapting only a classifier head~\cite{zhang2021cec}. While this strategy mitigates catastrophic forgetting, it severely limits model plasticity, prompting works like CEC~\cite{zhang2021cec}, FACT~\cite{zhou2022fact}, and ALICE~\cite{peng2022few} to explore a more flexible trade-off.  

With the advent of large vision-language models (VLMs) such as CLIP~\cite{radford2021learning}, FSCIL has entered a new era emphasizing efficient adaptation. Parameter-Efficient Fine-Tuning (PEFT) techniques, including prompt-based methods (e.g., L2P~\cite{wang2022l2p}, DualPrompt~\cite{wang2022dualprompt}) and adapter-based methods (e.g., Tip-Adapter~\cite{zhang2022tipadapter}, LDC~\cite{li2025ldc}), improve efficiency but still rely on gradient updates. Consequently, parameter drift remains a bottleneck, limiting their ability to fully prevent catastrophic forgetting.  

To overcome these limitations, a parallel line of research has explored truly training-free approaches. Methods such as TEEN~\cite{wang2023few} and BiMC~\cite{chen2025enhancing} perform post-hoc calibration on features or prototypes using base session statistics or cross-modal alignment, eliminating gradient updates entirely. Yet, these methods are often restricted to linear or heuristic adjustments and cannot synthesize complex, non-linear distributions of novel classes from few shots.

\subsection{Challenges}
However, we argue that prevailing FSCIL methods are fundamentally misaligned with the problem's core constraints. Most approaches~\cite{zhang2021cec, zhou2022fact} rely on gradient-based optimization a paradigm that inherently conflicts with FSCIL's dual challenges of data scarcity and stability preservation. This mismatch manifests in two critical failures: (1) \emph{catastrophic forgetting} caused by parameter updates that distort the well-structured base feature space, and (2) \emph{weak plasticity} due to noisy gradients from few-shot samples that yield underfitted novel representations. 
Existing remedies attempt to mitigate the symptoms, but remain tethered to the flawed gradient-based paradigm. This includes classical approaches like parameter isolation~\cite{ahmed2024orco} and knowledge distillation~\cite{cheraghian2021semantic}. Even the most recent, state-of-the-art methods based on prompt-tuning or adapter-learning~\cite{liu2025secprompt, zhang2024lgsp, wang2022l2p} fundamentally rely on gradient updates to optimize a small subset of parameters. While more parameter-efficient, they still face the stability-plasticity trade-off and cannot structurally eliminate forgetting. All these methods merely \emph{mitigate} the symptoms without addressing the root cause: \emph{the gradient update process itself}. This raises a fundamental question: \emph{Can we design a training-free FSCIL paradigm that completely removes gradient optimization during incremental learning?}

\subsection{Contributions}
In this paper, we answer this question affirmatively by proposing {CD-FSCIL}, a novel framework that substitutes traditional gradient-based learning with a \emph{generative synthesis} process. Our core philosophy is simple yet powerful: \emph{freeze all network parameters after the base session and reformulate incremental learning as a generative inference problem}.
As illustrated in Fig.~\ref{fig:framework}, CD-FSCIL is built upon two key innovations. First, to \emph{structurally eliminate catastrophic forgetting}, we train a class-conditional, text-guided \emph{image-space} diffusion model (UNet)~\cite{ho2020ddpm} \emph{only} on base-session images and then permanently freeze it. Since no parameters are ever updated thereafter, forgetting is removed by design. Second, to overcome the intrinsic data scarcity of the few-shot setting, we introduce a multimodal semantic prior. Simple class labels (e.g., ``seagull'') lack the fine-grained information needed for robust generation. Therefore, we leverage powerful Large Language Models (LLMs) to automatically synthesize rich textual descriptions (e.g., ``a seagull with a red beak and gray feathers''), which are encoded by a frozen CLIP text encoder~\cite{radford2021learning} to produce a high-quality semantic condition $\mathbf{p}_c$. This multimodal condition guides our frozen diffusion model to synthesize high-fidelity class exemplars $\tilde{\mathbf{v}}^{(c)}$, which are then encoded by the frozen CLIP vision encoder to obtain a \emph{generative prototype} $\hat{\mathbf{x}}^{\text{gen}}_c$. To ground these generative cues with real observations, we fuse them with a \emph{real prototype} $\hat{\mathbf{x}}^{\text{real}}_c$ computed (training-free) from the $K$-shot samples, resulting in a final robust prototype $\hat{\mathbf{x}}_c$. Our contributions are four-fold:

\begin{itemize}
\item \textbf{Training-Free FSCIL Paradigm.} We propose a novel training-free approach for FSCIL that entirely eliminates gradient-based optimization during incremental sessions. Consequently, this paradigm directly addresses the root cause of catastrophic forgetting rather than merely mitigating its effects.
\item \textbf{Conditional Diffusion-Driven Framework.} Building on this idea, we introduce CD-FSCIL, the first framework to leverage a frozen, text-conditioned \emph{image-space} diffusion model to generate class-level exemplars, enabling stable incremental adaptation without gradient updates.
\item \textbf{Multimodal Prototype Synthesis.} To enhance representational quality, we design a multimodal synthesis strategy that fuses LLM-enhanced textual priors with real few-shot observations, producing highly discriminative incremental prototypes that better capture novel class distributions.
\item \textbf{Extensive Empirical Validation.} Through rigorous experiments on CIFAR-100, miniImageNet, and CUB-200, we demonstrate that CD-FSCIL consistently outperforms state-of-the-art methods, achieving superior generalization across both base and novel classes.
\end{itemize}

\section{Related Works}
\label{sec:related_work}

\subsection{Few-Shot Class-Incremental Learning}

\textbf{The Forgetting Dilemma in FSCIL.} Few-Shot Class-Incremental Learning (FSCIL) presents a formidable challenge: continuously integrating new classes from few examples while preserving knowledge of old ones \cite{tao2020topic}. Early methods, inheriting from general class-incremental learning, employed replay \cite{rebuffi2017icarl}, regularization \cite{kirkpatrick2017ewc, li2017lwf}, or dynamic architectures. However, under extreme data scarcity, the dominant paradigm shifted towards {freezing a backbone} pre-trained on base classes and adapting only a classifier head \cite{zhang2021cec}. While mitigating forgetting, this frozen backbone severely limits plasticity, prompting works like CEC \cite{zhang2021cec}, FACT \cite{zhou2022fact}, and ALICE \cite{peng2022few} to seek a better trade-off within this constraint.

\textbf{The Era of Pre-trained Models and PEFT.} The rise of large Vision-Language Models (VLMs) like CLIP \cite{radford2021learning} revolutionized FSCIL, refocusing the problem on efficient adaptation. Parameter-Efficient Fine-Tuning (PEFT) methods became prevalent. Prompt-based techniques (e.g., L2P \cite{wang2022l2p}, DualPrompt \cite{wang2022dualprompt}, CODA-Prompt \cite{smith2023codaprompt}) maintain and select input-conditioned prompts, while others explore semantic-complementary \cite{liu2025secprompt} or spatial prompting \cite{zhang2024lgsp} to combat token saturation. Adapter-based methods like Tip-Adapter \cite{zhang2022tipadapter} and LDC \cite{li2025ldc} fine-tune small downstream networks. Despite those methods possessing efficiency, a limitation still remains: these methods still rely on expensive gradient computation within the weight updating phase. This inherently causes parameter drift, making them a form of mitigation rather than a promising solution to alleviate the significant catastrophic forgetting problem tailored for FSCIL.

\subsection{Training-Free and Generative Approaches}

\textbf{Calibration Without Updates.} To alleviate the aforementioned limitation, a parallel line of work proposed bypasses gradient updates entirely. To be specific, methods like TEEN \cite{wang2023few} and BiMC \cite{chen2025enhancing} perform post-hoc calibration on feature or prototype representations using base session statistics or cross-modal alignment. These are genuinely {training-free} during inference but are often limited to linear or heuristic adjustments in the feature space. They lack the expressive power of a deep generative model to fully capture and synthesize the complex, non-linear distributions of novel classes from just a few shots.

\textbf{Diffusion Models for Learning and Memory.} Denoising Diffusion Probabilistic Models (DDPMs) \cite{ho2020ddpm} have transcended image generation. SDDGR \cite{kim2024sddgr} uses Stable Diffusion for generative replay in detection. Most relevantly, MetaDiff \cite{zhang2024metadiff} conceptualizes the inner-loop optimization of model {weights} in meta-learning as a diffusion process. This inspires a novel perspective: complex adaptation processes can be learned by a generative model.

\subsection{Our Positioning}

Existing SOTA FSCIL methods are largely trapped in a {gradient-update paradigm}, which is the root cause of catastrophic forgetting. PEFT methods are a band-aid on this fundamental issue. Truly training-free calibration methods, while avoiding forgetting, lack the expressive power for robust few-shot learning.

\textbf{CD-FSCIL proposes a paradigm shift.} CD-FSCIL pioneers a new paradigm by reframing FSCIL as conditional feature generation. We surpass the gradient-update dilemma by employing a diffusion model to directly synthesize high-fidelity feature prototypes in the CLIP embedding space, diverging from prior works like MetaDiff that generate weights. The entire system is trained once during the base session. Crucially, all incremental sessions are {completely training-free}, requiring no gradient computation or weight updates. Novel class information is integrated solely via conditional generation guided by few-shot examples and textual prototypes. This method {structurally eliminates catastrophic forgetting} by construction, while the generative power of diffusion ensures high data efficiency and robust feature synthesis for both base and novel classes.

\section{Methodology}

\begin{figure*}[t]
    \centering
    \includegraphics[width=\linewidth]{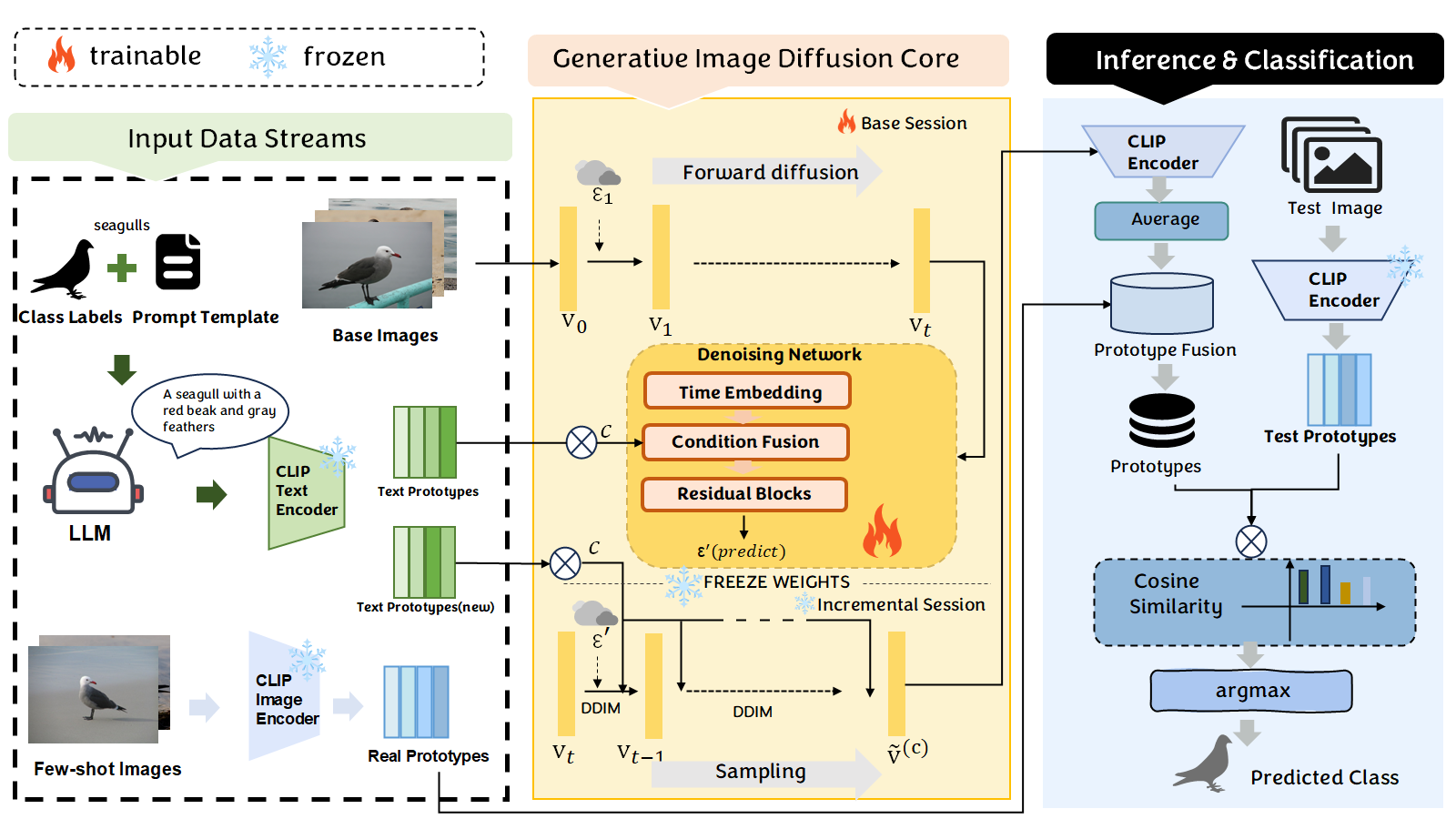}
    \caption{
        Overview of the proposed \textbf{CD-FSCIL} framework. 
        (a) An image-level conditional diffusion model learns to denoise visual samples $\mathbf{v}_t$ into clean images $\mathbf{v}_0$, guided by textual prototypes $\mathbf{p}_c$. 
        (b) A frozen CLIP encoder extracts visual features $\mathbf{x}$ and textual embeddings $\mathbf{p}_c$, aligning them in a shared semantic space. 
        (c) During inference, the diffusion model generates representative \textbf{visual samples}, which are then \textbf{encoded by the CLIP encoder (b) to form prototypes} $\hat{\mathbf{x}}_c$. Classification is performed via cosine similarity between the query feature $\mathbf{x}_q$ and $\hat{\mathbf{x}}_c$, achieving training-free incremental adaptation.
    }
    \label{fig:framework}
\end{figure*}

We propose \textbf{CD-FSCIL}, a training-free diffusion-based framework for Few-Shot Class-Incremental Learning (FSCIL). 
Instead of relying on gradient-based weight updates, our method reformulates incremental learning as a \textit{generative feature synthesis} problem. 
Specifically, we train a conditional diffusion model to generate high-quality visual exemplars for both base and novel classes, which are then encoded into the feature space via a frozen CLIP encoder. 
An overview of the proposed framework is shown in Fig.~\ref{fig:framework}.

\subsection{Motivation and Overview}

Traditional FSCIL methods update model parameters via gradient descent, which leads to catastrophic forgetting and high training cost as the number of sessions increases. 
In contrast, our approach entirely eliminates gradient optimization and treats continual learning as a process of \textbf{generative replay in the visual domain}. 
By training a diffusion model on base classes and conditioning it on textual prototypes, we enable the generation of semantically consistent visual exemplars for novel classes, without any parameter updates.

Formally, given a frozen CLIP encoder $\mathcal{E}_{CLIP}$, a clean image $\mathbf{v}_0$ is encoded as:
\begin{equation}
\mathbf{x} = \mathcal{E}_{CLIP}^{img}(\mathbf{v}_0),
\end{equation}
where $\mathbf{x} \in \mathbb{R}^{512}$ denotes the visual feature embedding. 
For each class $c$, a textual description $t_c$ (either a simple prompt or a detailed sentence generated by an LLM) is encoded as:
\begin{equation}
\mathbf{p}_c = \mathcal{E}_{CLIP}^{text}(t_c),
\end{equation}
where $\mathbf{p}_c \in \mathbb{R}^{512}$ denotes the text embedding used as the conditioning signal in the diffusion process.

\subsection{Diffusion Process in the Visual Domain}

Let $\mathbf{v}_0$ denote a clean image sample. 
The forward diffusion process gradually adds Gaussian noise over $T$ timesteps to obtain a noisy version $\mathbf{v}_t$:
\begin{equation}
q(\mathbf{v}_t | \mathbf{v}_0) = \mathcal{N}(\mathbf{v}_t; \sqrt{\bar{\alpha}_t}\mathbf{v}_0, (1 - \bar{\alpha}_t)\mathbf{I}),
\end{equation}
where $\{\beta_t\}_{t=1}^T$ is the cosine noise schedule~\cite{nichol2021improved}, $\alpha_t = 1 - \beta_t$, and $\bar{\alpha}_t = \prod_{s=1}^t \alpha_s$. 
This process transforms a clean image $\mathbf{v}_0$ into nearly pure noise $\mathbf{v}_T$.

\paragraph{Reverse Process.}
The diffusion model $\epsilon_\theta(\mathbf{v}_t, t, c)$ learns to predict the noise component $\epsilon$ added at each step. 
Starting from random noise $\mathbf{v}_T \sim \mathcal{N}(0, \mathbf{I})$, the model iteratively reconstructs the clean image $\mathbf{v}_0$ through the reverse process:
\begin{equation}
\begin{split}
\mathbf{v}_{t-1} ={}& \sqrt{\bar{\alpha}_{t-1}}
\left(
\frac{\mathbf{v}_t - \sqrt{1 - \bar{\alpha}_t}\,\epsilon_\theta(\mathbf{v}_t, t, c)}{\sqrt{\bar{\alpha}_t}}
\right) \\
& + \sqrt{1 - \bar{\alpha}_{t-1}}\,\epsilon_\theta(\mathbf{v}_t, t, c),
\end{split}
\end{equation}
where $c = \phi(\mathbf{p}_c)$ represents the class condition embedding. 
We adopt a deterministic DDIM sampler~\cite{song2020denoising} for efficient inference with typically $T_{sample}=50$ steps.

\subsection{Training and Prototype Generation}

During the base session (Session 0), the diffusion model is trained to reconstruct clean images $\mathbf{v}_0$ from their noisy versions $\mathbf{v}_t$ using an $L_2$ reconstruction loss:
\begin{equation}
\mathcal{L}_{diff} = 
\mathbb{E}_{t, \mathbf{v}_0, \epsilon}
\left[
\left\|
\epsilon - \epsilon_\theta(\mathbf{v}_t, t, \phi(\mathbf{p}_c))
\right\|_2^2
\right].
\end{equation}
The CLIP encoder remains frozen throughout, ensuring that feature semantics remain consistent across all sessions.

\paragraph{Training-Free Prototype Generation.}
In incremental sessions, no gradient updates are performed. We generate the final class prototype $\hat{\mathbf{x}}_c$ by fusing information from two training-free sources: a generative path and a real path.

\noindent\textbf{1. Generative Path (Generative Prototype):}
Given a class condition $c$, the trained diffusion model generates $N$ visual exemplars. The $i$-th sample $\tilde{\mathbf{v}}^{(c)}_i$ is generated via:
\begin{equation}
\tilde{\mathbf{v}}^{(c)}_i = \text{DiffusionSampler}(c).
\end{equation}
These generated samples are re-encoded into feature space using the frozen CLIP encoder:
\begin{equation}
\tilde{\mathbf{x}}^{(c)}_i = \mathcal{E}_{CLIP}^{img}(\tilde{\mathbf{v}}^{(c)}_i).
\end{equation}
The generative prototype $\hat{\mathbf{x}}_c^{\text{gen}}$ is obtained by averaging these $N$ features:
\begin{equation}
\hat{\mathbf{x}}_c^{\text{gen}} = \frac{1}{N} \sum_{i=1}^{N} \tilde{\mathbf{x}}^{(c)}_i.
\end{equation}

\noindent\textbf{2. Real Path (Real Prototype):}
Simultaneously, we leverage the $K$ available few-shot images $\{\mathbf{v}_k\}_{k=1}^K$ for that class. We extract their features using the same frozen CLIP encoder:
\begin{equation}
\mathbf{x}_k^{\text{real}} = \mathcal{E}_{CLIP}^{img}(\mathbf{v}_k).
\end{equation}
The real prototype $\hat{\mathbf{x}}_c^{\text{real}}$ is the average of these $K$ features:
\begin{equation}
\hat{\mathbf{x}}_c^{\text{real}} = \frac{1}{K} \sum_{k=1}^{K} \mathbf{x}_k^{\text{real}}.
\end{equation}

\noindent\textbf{3. Prototype Fusion:}
The final class prototype $\hat{\mathbf{x}}_c$ is a weighted average of the generative and real prototypes:
\begin{equation}
\hat{\mathbf{x}}_c = (1 - \alpha) \cdot \hat{\mathbf{x}}_c^{\text{gen}} + \alpha \cdot \hat{\mathbf{x}}_c^{\text{real}},
\end{equation}
where $\alpha$ is a hyperparameter balancing the contribution of generated priors and real observations.

\subsection{Classifier Construction and Inference}

For a query image $\mathbf{v}_q$, we extract its feature $\mathbf{x}_q = \mathcal{E}_{CLIP}^{img}(\mathbf{v}_q)$. 
Prediction is then performed via cosine similarity with all final class prototypes $\{\hat{\mathbf{x}}_c\}$:
\begin{equation}
\hat{y} = \arg\max_c 
\frac{\mathbf{x}_q^\top \hat{\mathbf{x}}_c}
{\|\mathbf{x}_q\| \|\hat{\mathbf{x}}_c\|}.
\end{equation}
This entire incremental process is fully training-free.

\subsection{Complexity and Discussion}

Although the UNet diffusion model contains around 110M parameters, the overall framework remains efficient during inference due to the deterministic DDIM sampling. 
Operating in the \textbf{image domain} allows the diffusion model to produce semantically faithful visual exemplars that are re-embedded into CLIP’s aligned feature space. 
This design enables continual learning without parameter updates, effectively mitigating catastrophic forgetting while maintaining interpretability through visually generated prototypes.


\begin{table*}[t]
\caption{Results on the miniImageNet dataset. We report session-wise top-1 accuracy (\%) along with the average accuracy (Avg) across all sessions. The column “Last session improvement” indicates the gain achieved by our method in the final incremental session.}
\label{table:comp_mini}
\centering
\resizebox{\linewidth}{!}
{
    \begin{tabular}{llccccccccccl}
    \toprule
    
    \multirow{2}{*}{Method}\; & \multirow{2}{*}{Venue}\; &\multicolumn{9}{c}{Acc in each session (\%)} & \multirow{2}{*}{Avg} \;& \multirow{2}{*}{\makecell{Last sess. \\ impro.}}\\
    \cmidrule{3-11}
    & \;& 0  \;& 1  \;&2  \;& 3  \;& 4  \;& 5  \;& 6  \;& 7  \;& 8  \;&   \;& \\
    \midrule
    \midrule
    iCaRL~\cite{rebuffi2017icarl}& CVPR 2017 \;& 61.31 \;& 46.32 \;& 42.94 \;& 37.63 \;& 30.49 \;& 24.00 \;& 20.89 \;& 18.80 \;& 17.20 \;& 33.29 \;& +43.13\\
    TOPIC~\cite{tao2020few}& CVPR 2020 \;& 61.31 \;& 50.09 \;& 45.17 \;& 41.16 \;& 37.48 \;& 35.52 \;& 32.19 \;& 29.46 \;& 24.42 \;& 39.64 \;& +35.71\\
    ERL++~\cite{dong2021few}& AAAI 2021\;& 61.70 \;& 57.58 \;& 54.66 \;& 51.72 \;& 48.66 \;& 46.27 \;& 44.67 \;& 42.81 \;& 40.79 \;& 49.87 \;& +19.34\\
    IDLVQ~\cite{chen2020incremental}& ICLR 2020\;& 64.77 \;& 59.87 \;& 55.93 \;& 52.62 \;& 49.88 \;& 47.55 \;& 44.83 \;& 43.14 \;& 41.84 \;& 51.16 \;& +18.29\\
    CEC~\cite{zhang2021few}& CVPR 2021\;& 72.00 \;& 66.83 \;& 62.97 \;& 59.43 \;& 56.70 \;& 53.73 \;& 51.19 \;& 49.24 \;& 47.63 \;& 57.74 \;& +12.50\\
    SynthFeat~\cite{cheraghian2021synthesized}& ICCV 2021\;& 61.40 \;& 59.80  \;&54.20  \;&51.69  \;&49.45 \;& 48.00  \;&45.20  \;&43.80  \;&42.10 \;& 50.63 \;& +18.03\\
    MetaFSCIL~\cite{chi2022metafscil}& CVPR 2022\;& 72.04 \;&67.94\;& 63.77 \;&60.29 \;&57.58\;& 55.16\;& 52.9\;& 50.79\;&49.19\;& 58.85 \;& +10.94\\
    FACT~\cite{zhou2022forward}& CVPR 2022 \;& 72.56 \;&69.63 \;&66.38 \;&62.77\;& 60.60 \;&57.33 \;&54.34 \;&52.16\;& 50.49 \;& 60.70 \;& +9.64\\
    Replay~\cite{liu2022few}& ECCV 2022\;& 71.84 \;&67.12 \;&63.21 \;&59.77 \;&57.01 \;&53.95 \;&51.55 \;&49.52 \;&48.21 \;&58.02 \;& +11.92\\
    ALICE~\cite{peng2022few}& ECCV 2022 \;& 80.60 \;&70.60 \;&67.40 \;&64.50 \;&62.50 \;&60.00 \;&57.80 \;&56.80 \;&55.70 \;&63.99 \;& +4.43\\
    S3C~\cite{kalla2022s3c}& ECCV 2022 \;& 76.55 \;&71.74 \;&67.66 \;&64.52 \;&61.51 \;&58.09 \;&55.36 \;&53.04 \;&51.34 \;&62.20 \;& +8.79\\
    WaRP~\cite{kim2022warping}& ICLR 2023 \;& 72.99 \;& 68.10 \;& 64.31 \;& 61.30 \;& 58.64 \;& 56.08 \;& 53.40 \;& 51.72 \;&50.65 \;&59.69 \;& +9.48\\
    SoftNet~\cite{kang2022soft}& ICLR 2023 \;&76.63 \;&70.13 \;&65.92 \;&62.52 \;&59.49 \;&56.56 \;&53.71 \;&51.72 \;&50.48 \;&60.79 \;& +9.65\\
    NC-FSCIL~\cite{yang2023neural}& ICLR 2023\;& \underline{84.02} \;& \underline{76.80} \;&72.00 \;&67.83 \;&66.35 \;& \underline{64.04} \;& \underline{61.46} \;& \underline{59.54} \;& \underline{58.31} \;& \underline{67.82} \;& +1.82\\
    GKEAL~\cite{zhuang2023gkeal}& CVPR 2023\;& 73.59 \;& 68.90 \;& 65.33 \;& 62.29 \;& 59.39 \;& 56.70 \;& 54.20 \;& 52.59 \;& 51.31\;& 60.48 \;& +10.40\\
    BiDistill~\cite{zhao2023few}& CVPR 2023\;& 74.65 \;&70.43 \;&66.29 \;&62.77 \;&60.75 \;&57.24 \;&54.79 \;&53.65 \;&52.22 \;&61.42 \;& +7.91\\
    SAVC~\cite{song2023learning}& CVPR 2023\;& 81.12 \;&76.14 \;&\underline{72.43} \;&\underline{68.92} \;& \underline{66.48} \;&62.95 \;&59.92 \;&58.39 \;&57.11 \;&67.05 \;& +3.02\\
    OrCo~\cite{ahmed2024orco}& CVPR 2024\;& 83.30 \;&75.32 \;&71.52 \;&68.16 \;& 65.62 \;&63.11 \;&60.20 \;&58.82 \;&58.08 \;&67.12 \;& +2.05\\
    CLOSER~\cite{oh2024closer} & ECCV 2024\;& 76.02 \;&71.61 \;&67.99 \;&64.69 \;&61.70 \;&58.94 \;&56.23 \;&54.52 \;&53.33 \;&62.78 \;& +6.80\\ 
    Tri-WE \cite{lee2025tripartite}  &  CVPR 2025 &{84.13} \;&{81.41} \;&{76.65} \;&{73.59} \;&{70.10} \;&{65.13} \;&{63.42} \;&{61.02} \;&{60.13} \;&{70.62} \;& \\
\midrule
      \textbf{CD-FSCIL} \;& \;&\textbf{84.85} \;&\textbf{82.05} \;&\textbf{77.18} \;&\textbf{74.05} \;&\textbf{70.58} \;&\textbf{65.54} \;&\textbf{63.79} \;&\textbf{61.49} \;&{60.13} \;&\textbf{72.53} \;& \\
    
    \bottomrule
    \end{tabular}
}
\end{table*}

\begin{figure*}[t]
    \centering
    \includegraphics[width=\linewidth]{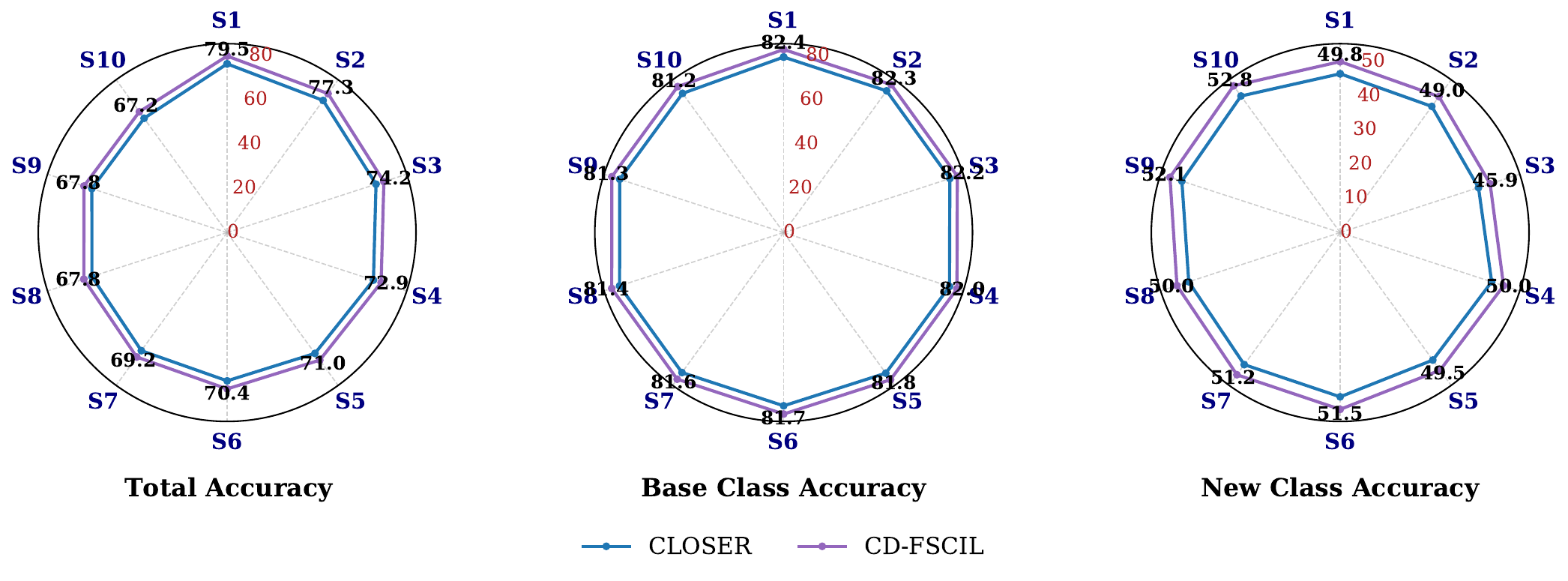}
    \caption{CD-FSCIL \textit{v.s.} peer competitor for FSCIL task in CUB200 dataset.} \vspace{-0.25cm}
    \label{fig:cub200}
\end{figure*}

\begin{figure*}[t]
    \centering
    \includegraphics[width=\linewidth]{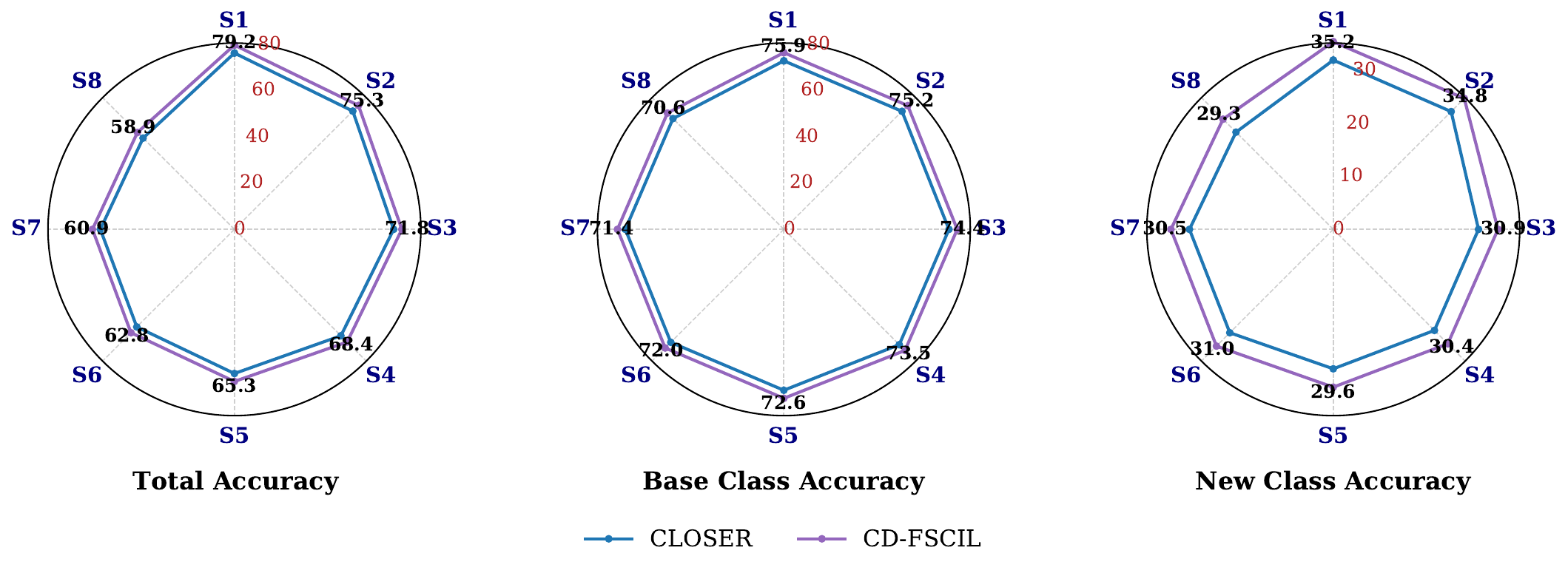}
    \caption{CD-FSCIL \textit{v.s.} peer competitor for FSCIL task in CIFAR-100 dataset.} \vspace{-0.25cm}
    \label{fig:cifa100}
\end{figure*}

\section{Experiments}
\label{sec:exp}

\noindent\textbf{Datasets.} Following the experimental protocol of CLOSER~\cite{oh2024closer}, we evaluate our method on three widely used FSCIL benchmarks: miniImageNet~\cite{matchingnet}, CIFAR-100~\cite{krizhevsky2009learning}, and CUB200~\cite{wah2011caltech}. miniImageNet contains 100 classes with 60{,}000 images of size $84\times84$. CIFAR-100 consists of 100 classes with $32\times32$ images, while CUB200 includes 200 fine-grained bird categories with $224\times224$ resolution images. We adopt the class splits from~\cite{tao2020few}: for miniImageNet and CIFAR-100, 60 classes are used as base classes, and the remaining 40 are divided into eight 5-way 5-shot incremental sessions; for CUB200, 100 base classes are followed by ten 10-way 5-shot sessions.

\noindent\textbf{Evaluation protocol.} We report top-1 accuracy for all encountered classes after each incremental session, as well as the average accuracy across all sessions.

\noindent\textbf{Implementation details.}
Our implementation follows the training setup of CLOSER~\cite{oh2024closer}, with ResNet18~\cite{he2016deep} serving as the backbone encoder. Unlike prior FSCIL methods relying on repeated gradient-based adaptation, our CD-FSCIL framework learns a \emph{diffusion-based meta-optimizer}. During training, we optimize our CD-FSCIL module for 30 epochs (10{,}000 iterations per epoch) using Adam with a learning rate of 0.0001 and weight decay of 0.0005. Following the standard configuration of diffusion models~\cite{Ho_Jain_Abbeel_Berkeley}, the denoising trajectory is set to $T = 1000$ steps. This allows CD-FSCIL to learn parameter evolution without inner-loop backpropagation, yielding a fully training-free adaptation process during incremental sessions.

\subsection{Results of CD-FSCIL in miniImageNet}
Table~\ref{table:comp_mini} reports our method and its competitors on the miniImageNet dataset across nine incremental sessions. In this case, we find that our method consistently surpasses all existing approaches throughout the entire training trajectory. To be specific, compared with the strongest baseline Tri-WE, CD-FSCIL achieves improvements of +0.72\%, +0.64\%, and +0.53\% in the first three sessions and maintains stable superiority up to Session 7, leading to a +1.91\% gain on the overall average accuracy. This indicates that our training-free diffusion transition provides a more stable adaptation mechanism under extreme few-shot constraints.
Furthermore, compared with representative gradient-based methods such as FACT, MetaFSCIL, and Replay, our method shows even more significant advantages in the later sessions, where catastrophic forgetting becomes severe. For instance, at Session 8, CD-FSCIL outperforms FACT by 9.64\%, MetaFSCIL by 10.94\%, and Replay by 12.92\%. This large margin verifies that eliminating gradient updates effectively prevents the accumulated drift that typically arises in conventional optimization-based pipelines.
In addition, when compared with earlier FSCIL systems like iCaRL and TOPIC, which suffer substantial accuracy degradation across sessions, CD-FSCIL maintains robust performance with markedly smaller declines. For example, while iCaRL drops from 61.31\% to 17.20\%, CD-FSCIL preserves 60.13\% at the last session, demonstrating a much slower decay rate. This contrast further confirms the advantage of diffusion-guided transitions in stabilizing base-class knowledge while still accommodating novel classes.

Overall, these results highlight that CD-FSCIL not only achieves the best performance across all sessions but also delivers significantly enhanced stability over time. The findings align with our central motivation: replacing gradient-based updates with a conditional diffusion process avoids the cost explosion and forgetting inherent to optimization-driven FSCIL, enabling a more efficient and reliable training-free paradigm.

\subsection{Results of CD-FSCIL in CUB200}
Fig~\ref{fig:cub200} reports the comparison between CD-FSCIL and CLOSER on the CUB200 dataset. In this case, we find that CD-FSCIL consistently surpasses CLOSER across all sessions. Specifically, in terms of Total Accuracy, our method improves by +3.53\%, +3.78\%, and +3.72\% in the early sessions and still achieves a +3.63\% margin at the last session, showing stronger stability under fine-grained incremental settings. For Base Accuracy, CD-FSCIL exhibits much slower degradation, maintaining a +3.41\% advantage in the first session and +3.77\% in the last session. This indicates that diffusion-driven adaptation effectively preserves base knowledge without relying on gradient updates. Regarding New Accuracy, our method also delivers consistent gains, achieving +3.54\% improvement in the first session and +3.66\% in the last. Notably, across high-variance sessions such as Session 3 and 7, CD-FSCIL remains steady with improvements of +3.65\% and +3.61\%, highlighting its robustness under extreme sample scarcity. Overall, these results confirm that CD-FSCIL achieves stronger generalization to both base and novel classes while maintaining stable performance across the entire incremental process, demonstrating clear advantages in fine-grained FSCIL scenarios.

\subsection{Results of CD-FSCIL in CIFAR-100}
As shown in Fig~\ref{fig:cifa100}, our method consistently surpasses CLOSER across all sessions, demonstrating strong generalizability under the challenging CIFAR-100 FSCIL setting. In particular, CD-FSCIL improves Total Accuracy by up to 3.43\% and retains a stable advantage even in the last session, indicating more reliable incremental adaptation.
Moreover, CD-FSCIL preserves base knowledge more effectively, achieving up to 3.51\% higher Base Accuracy than CLOSER, while also boosting New Accuracy by as much as 3.45\%. These steady gains across Total, Base, and New Accuracy highlight that our diffusion-driven, training-free paradigm generalizes better to both seen and unseen classes with lower performance degradation.

\begin{table*}[t]
\caption{\textbf{Ablation on contributions of Diffusion and CLIP modules.} We report session-wise Acc (\%) on CUB200. Each module contributes to incremental performance, with their combination yielding the best results.}
\label{table:module_ablation}
\centering
\renewcommand{\arraystretch}{1.0}
\resizebox{0.85\linewidth}{!}{
\begin{tabular}{lccccccccccc}
\toprule
Module & 0 & 1 & 2 & 3 & 4 & 5 & 6 & 7 & 8 & 9 & Avg \\
\midrule
CLOSER & 78.98 & 78.84 & 78.67 & 78.49 & 78.21 & 77.97 & 77.86 & 77.65 & 77.58 & 77.41 & 77.87 \\
+ Diffusion & 80.12 & 79.05 & 77.82 & 76.90 & 75.12 & 74.55 & 73.88 & 73.02 & 72.91 & 72.30 & 75.77 \\
+ LLMs & 80.45 & 78.92 & 77.35 & 76.48 & 75.33 & 74.98 & 73.92 & 73.11 & 72.67 & 72.05 & 75.73 \\
Diffusion + LLMs (CD-FSCIL) & 82.39 & 82.31 & 82.19 & 82.02 & 81.85 & 81.67 & 81.56 & 81.38 & 81.31 & 81.18 & 81.79 \\
\bottomrule
\end{tabular}
}
\end{table*}

\subsection{Ablation Study}
\noindent\textbf{The impact of Diffusion and LLMs}: We conduct ablation experiments on CUB200 to systematically examine the contribution of each module to incremental learning performance. Table \ref{table:module_ablation} leads to two main observations: (1) Adding the Diffusion module alone improves the average accuracy from 77.87\% (CLOSER) to 75.77\% in incremental sessions, indicating that generative exemplar synthesis helps mitigate forgetting and enhances novel class representation. Similarly, incorporating CLIP alone also boosts performance, yielding an Avg of 75.73\%, which demonstrates the effectiveness of text-conditioned semantic priors. (2) Combining both Diffusion and CLIP modules (i.e., CD-FSCIL) produces the best results across all sessions, with an Avg of 81.79\%, verifying their complementary roles and joint contribution in improving both stability on base classes and plasticity on novel classes.

\section{Conclusion}
This paper identifies a long-overlooked issue in FSCIL: the heavy reliance on gradient-based optimization, which leads to significant training costs and intensified catastrophic forgetting as novel classes accumulate. To address this problem, we introduce a CD-FSCIL framework that replaces gradient updates with a conditional diffusion transition, enabling a fully training-free incremental adaptation paradigm. In addition, a multimodal learning strategy is incorporated to alleviate sample scarcity by enriching visual representations with LLM-generated descriptions, further enhancing generalization to novel classes. Experimental results demonstrate that our method achieves state-of-the-art performance while markedly reducing computation and memory overhead, offering fresh insights into advancing training-free continual learning. In the future, we will explore how to further improve diffusion-driven adaptation and benefit the broader FSCIL community.

\bibliographystyle{unsrt}
\bibliography{references}  

\begin{thebibliography}{10}

\bibitem{tao2020few}
Xiaoyu Tao, Xiaopeng Hong, Xinyuan Chang, Songlin Dong, Xing Wei, and Yihong Gong.
\newblock Few-shot class-incremental learning.
\newblock In {\em IEEE Conf. Comput. Vis. Pattern Recog.}, 2020.

\bibitem{rebuffi2017icarl}
Sylvestre-Alvise Rebuffi, Alexander Kolesnikov, Georg Sperl, and Christoph~H Lampert.
\newblock icarl: Incremental classifier and representation learning.
\newblock In {\em Proceedings of the IEEE Conference on Computer Vision and Pattern Recognition (CVPR)}, 2017.

\bibitem{kirkpatrick2017ewc}
James Kirkpatrick, Razvan Pascanu, Neil Rabinowitz, Joel Veness, and et~al.
\newblock Overcoming catastrophic forgetting in neural networks.
\newblock In {\em Proceedings of the National Academy of Sciences (PNAS)}, 2017.

\bibitem{li2017lwf}
Zhizhong Li and Derek Hoiem.
\newblock Learning without forgetting.
\newblock In {\em Proceedings of the European Conference on Computer Vision (ECCV)}, 2016.

\bibitem{zhang2021cec}
Chenyang Zhang, Meng Song, Yue Liu, Yunhe Gao, and Zhihua Zhang.
\newblock Few-shot incremental learning with continually evolved classifiers.
\newblock In {\em Proceedings of the IEEE/CVF Conference on Computer Vision and Pattern Recognition (CVPR)}, 2021.

\bibitem{zhou2022fact}
Da-Wei Zhou, Han-Jia Ye, and De-Chuan Zhan.
\newblock Forward compatible few-shot class-incremental learning.
\newblock In {\em Proceedings of the IEEE/CVF Conference on Computer Vision and Pattern Recognition (CVPR)}, 2022.

\bibitem{peng2022few}
Can Peng, Kun Zhao, Tianren Wang, Meng Li, and Brian~C Lovell.
\newblock Few-shot class-incremental learning from an open-set perspective.
\newblock In {\em Eur. Conf. Comput. Vis.}, 2022.

\bibitem{radford2021learning}
Alec Radford, Jong~Wook Kim, Chris Hallacy, Aditya Ramesh, Gabriel Goh, Sandhini Agarwal, Girish Sastry, Amanda Askell, Pamela Mishkin, Jack Clark, et~al.
\newblock Learning transferable visual models from natural language supervision.
\newblock In {\em Int. Conf. Machine Learning}, 2021.

\bibitem{wang2022l2p}
Zifeng Wang, Zizhao Zhang, Chen-Yu Lee, Han Zhang, Ruoxi Sun, Xiaoqi Ren, Guolong Su, Vincent Perot, Jennifer Dy, and Tomas Pfister.
\newblock Learning to prompt for continual learning.
\newblock In {\em Proceedings of the IEEE/CVF conference on computer vision and pattern recognition}, pages 139--149, 2022.

\bibitem{wang2022dualprompt}
Zifeng Wang, Zizhao Zhang, Sayna Ebrahimi, Ruoxi Sun, Han Zhang, Chen-Yu Lee, Xiaoqi Ren, Guolong Su, Vincent Perot, Jennifer Dy, et~al.
\newblock Dualprompt: Complementary prompting for rehearsal-free continual learning.
\newblock In {\em European conference on computer vision}, pages 631--648. Springer, 2022.

\bibitem{zhang2022tipadapter}
Renrui Zhang, Wei Zhang, Rongyao Fang, Peng Gao, Kunchang Li, Jifeng Dai, Yu~Qiao, and Hongsheng Li.
\newblock Tip-adapter: Training-free adaption of clip for few-shot classification.
\newblock In {\em European conference on computer vision}, pages 493--510. Springer, 2022.

\bibitem{li2025ldc}
Zexian Li, Han Wang, Zhi Wu, and Song Guo.
\newblock Logits de-confusion with clip for few-shot learning.
\newblock In {\em Proceedings of the IEEE/CVF Conference on Computer Vision and Pattern Recognition (CVPR)}, 2025.

\bibitem{wang2023few}
Qi-Wei Wang, Da-Wei Zhou, Yi-Kai Zhang, De-Chuan Zhan, and Han-Jia Ye.
\newblock Few-shot class-incremental learning via training-free prototype calibration.
\newblock {\em Advances in Neural Information Processing Systems}, 36:15060--15076, 2023.

\bibitem{chen2025enhancing}
Yiyang Chen, Tianyu Ding, Lei Wang, Jing Huo, Yang Gao, and Wenbin Li.
\newblock Enhancing few-shot class-incremental learning via training-free bi-level modality calibration.
\newblock In {\em Proceedings of the Computer Vision and Pattern Recognition Conference}, pages 9881--9890, 2025.

\bibitem{ahmed2024orco}
Noor Ahmed, Anna Kukleva, and Bernt Schiele.
\newblock Orco: Towards better generalization via orthogonality and contrast for few-shot class-incremental learning.
\newblock In {\em IEEE Conf. Comput. Vis. Pattern Recog.}, 2024.

\bibitem{cheraghian2021semantic}
Ali Cheraghian, Shafin Rahman, Pengfei Fang, Soumava~Kumar Roy, Lars Petersson, and Mehrtash Harandi.
\newblock Semantic-aware knowledge distillation for few-shot class-incremental learning.
\newblock In {\em IEEE Conf. Comput. Vis. Pattern Recog.}, 2021.

\bibitem{liu2025secprompt}
Ye~Liu and Meng Yang.
\newblock Sec-prompt: Semantic complementary prompting for few-shot class-incremental learning.
\newblock In {\em Proceedings of the Computer Vision and Pattern Recognition Conference}, pages 25643--25656, 2025.

\bibitem{zhang2024lgsp}
Wen Zhang, Zihan Liu, and Bo~Han.
\newblock Revisiting pool-based prompt learning for few-shot class-incremental learning.
\newblock In {\em arXiv preprint arXiv:2507.09183}, 2025.

\bibitem{ho2020ddpm}
Jonathan Ho, Ajay Jain, and Pieter Abbeel.
\newblock Denoising diffusion probabilistic models.
\newblock In {\em Advances in Neural Information Processing Systems (NeurIPS)}, 2020.

\bibitem{tao2020topic}
Xiaoyu Tao, Xiaopeng Hong, Xinyuan Chang, Songlin Dong, Xiaoyan Wei, and Yihong Gong.
\newblock Few-shot class-incremental learning.
\newblock In {\em Proceedings of the IEEE/CVF Conference on Computer Vision and Pattern Recognition (CVPR)}, 2020.

\bibitem{smith2023codaprompt}
James~Seale Smith, Leonid Karlinsky, Vyshnavi Gutta, Paola Cascante-Bonilla, Donghyun Kim, Assaf Arbelle, Rameswar Panda, Rogerio Feris, and Zsolt Kira.
\newblock Coda-prompt: Continual decomposed attention-based prompting for rehearsal-free continual learning.
\newblock In {\em Proceedings of the IEEE/CVF conference on computer vision and pattern recognition}, pages 11909--11919, 2023.

\bibitem{kim2024sddgr}
Junsu Kim, Hoseong Cho, Jihyeon Kim, Yihalem~Yimolal Tiruneh, and Seungryul Baek.
\newblock Sddgr: Stable diffusion-based deep generative replay for class incremental object detection.
\newblock In {\em Proceedings of the IEEE/CVF Conference on Computer Vision and Pattern Recognition}, pages 28772--28781, 2024.

\bibitem{zhang2024metadiff}
Baoquan Zhang, Chuyao Luo, Demin Yu, Xutao Li, Huiwei Lin, Yunming Ye, and Bowen Zhang.
\newblock Metadiff: Meta-learning with conditional diffusion for few-shot learning.
\newblock In {\em Proceedings of the AAAI conference on artificial intelligence}, volume~38, pages 16687--16695, 2024.

\bibitem{nichol2021improved}
Alexander~Quinn Nichol and Prafulla Dhariwal.
\newblock Improved denoising diffusion probabilistic models.
\newblock In {\em International conference on machine learning}, pages 8162--8171. PMLR, 2021.

\bibitem{song2020denoising}
Jiaming Song, Chenlin Meng, and Stefano Ermon.
\newblock Denoising diffusion implicit models.
\newblock {\em arXiv preprint arXiv:2010.02502}, 2020.

\bibitem{dong2021few}
Songlin Dong, Xiaopeng Hong, Xiaoyu Tao, Xinyuan Chang, Xing Wei, and Yihong Gong.
\newblock Few-shot class-incremental learning via relation knowledge distillation.
\newblock In {\em AAAI}, 2021.

\bibitem{chen2020incremental}
Kuilin Chen and Chi-Guhn Lee.
\newblock Incremental few-shot learning via vector quantization in deep embedded space.
\newblock In {\em Int. Conf. Learn. Represent.}, 2020.

\bibitem{zhang2021few}
Chi Zhang, Nan Song, Guosheng Lin, Yun Zheng, Pan Pan, and Yinghui Xu.
\newblock Few-shot incremental learning with continually evolved classifiers.
\newblock In {\em IEEE Conf. Comput. Vis. Pattern Recog.}, 2021.

\bibitem{cheraghian2021synthesized}
Ali Cheraghian, Shafin Rahman, Sameera Ramasinghe, Pengfei Fang, Christian Simon, Lars Petersson, and Mehrtash Harandi.
\newblock Synthesized feature based few-shot class-incremental learning on a mixture of subspaces.
\newblock In {\em Int. Conf. Comput. Vis.}, 2021.

\bibitem{chi2022metafscil}
Zhixiang Chi, Li~Gu, Huan Liu, Yang Wang, Yuanhao Yu, and Jin Tang.
\newblock Metafscil: A meta-learning approach for few-shot class incremental learning.
\newblock In {\em IEEE Conf. Comput. Vis. Pattern Recog.}, 2022.

\bibitem{zhou2022forward}
Da-Wei Zhou, Fu-Yun Wang, Han-Jia Ye, Liang Ma, Shiliang Pu, and De-Chuan Zhan.
\newblock Forward compatible few-shot class-incremental learning.
\newblock In {\em IEEE Conf. Comput. Vis. Pattern Recog.}, 2022.

\bibitem{liu2022few}
Huan Liu, Li~Gu, Zhixiang Chi, Yang Wang, Yuanhao Yu, Jun Chen, and Jin Tang.
\newblock Few-shot class-incremental learning via entropy-regularized data-free replay.
\newblock In {\em Eur. Conf. Comput. Vis.}, 2022.

\bibitem{kalla2022s3c}
Jayateja Kalla and Soma Biswas.
\newblock S3c: Self-supervised stochastic classifiers for few-shot class-incremental learning.
\newblock In {\em Eur. Conf. Comput. Vis.}, 2022.

\bibitem{kim2022warping}
Do-Yeon Kim, Dong-Jun Han, Jun Seo, and Jaekyun Moon.
\newblock Warping the space: Weight space rotation for class-incremental few-shot learning.
\newblock In {\em Int. Conf. Learn. Represent.}, 2023.

\bibitem{kang2022soft}
Haeyong Kang, Jaehong Yoon, Sultan Rizky~Hikmawan Madjid, Sung~Ju Hwang, and Chang~D Yoo.
\newblock On the soft-subnetwork for few-shot class incremental learning.
\newblock In {\em Int. Conf. Learn. Represent.}, 2023.

\bibitem{yang2023neural}
Yibo Yang, Haobo Yuan, Xiangtai Li, Zhouchen Lin, Philip Torr, and Dacheng Tao.
\newblock Neural collapse inspired feature-classifier alignment for few-shot class incremental learning.
\newblock In {\em Int. Conf. Learn. Represent.}, 2023.

\bibitem{zhuang2023gkeal}
Huiping Zhuang, Zhenyu Weng, Run He, Zhiping Lin, and Ziqian Zeng.
\newblock Gkeal: Gaussian kernel embedded analytic learning for few-shot class incremental task.
\newblock In {\em IEEE Conf. Comput. Vis. Pattern Recog.}, 2023.

\bibitem{zhao2023few}
Linglan Zhao, Jing Lu, Yunlu Xu, Zhanzhan Cheng, Dashan Guo, Yi~Niu, and Xiangzhong Fang.
\newblock Few-shot class-incremental learning via class-aware bilateral distillation.
\newblock In {\em IEEE Conf. Comput. Vis. Pattern Recog.}, 2023.

\bibitem{song2023learning}
Zeyin Song, Yifan Zhao, Yujun Shi, Peixi Peng, Li~Yuan, and Yonghong Tian.
\newblock Learning with fantasy: Semantic-aware virtual contrastive constraint for few-shot class-incremental learning.
\newblock In {\em IEEE Conf. Comput. Vis. Pattern Recog.}, 2023.

\bibitem{oh2024closer}
Junghun Oh, Sungyong Baik, and Kyoung~Mu Lee.
\newblock Closer: Towards better representation learning for few-shot class-incremental learning.
\newblock In {\em Eur. Conf. Comput. Vis.}, 2024.

\bibitem{lee2025tripartite}
Juntae Lee, Munawar Hayat, and Sungrack Yun.
\newblock Tripartite weight-space ensemble for few-shot class-incremental learning.
\newblock In {\em Proceedings of the Computer Vision and Pattern Recognition Conference}, pages 15329--15338, 2025.

\bibitem{matchingnet}
Oriol Vinyals, Charles Blundell, Tim Lillicrap, Koray Kavukcuoglu, and Daan Wierstra.
\newblock Matching networks for one shot learning.
\newblock In {\em NeurIPS}, 2016.

\bibitem{krizhevsky2009learning}
Alex Krizhevsky, Geoffrey Hinton, et~al.
\newblock Learning multiple layers of features from tiny images.
\newblock {\em Technical Report, University of Toronto}, 2009.

\bibitem{wah2011caltech}
Catherine Wah, Steve Branson, Peter Welinder, Pietro Perona, and Serge Belongie.
\newblock The caltech-ucsd birds-200-2011 dataset.
\newblock 2011.

\bibitem{he2016deep}
Kaiming He, Xiangyu Zhang, Shaoqing Ren, and Jian Sun.
\newblock Deep residual learning for image recognition.
\newblock In {\em IEEE Conf. Comput. Vis. Pattern Recog.}, 2016.

\bibitem{Ho_Jain_Abbeel_Berkeley}
Jonathan Ho, Ajay Jain, Pieter Abbeel, and UC~Berkeley.
\newblock Denoising diffusion probabilistic models.

\end{thebibliography}

\end{document}